\documentclass{article} 
\usepackage{iclr2023_conference,times}

\usepackage{amsmath,amsfonts,bm}

\def\eqref#1{equation~\ref{#1}}

\def\1{\bm{1}}

\DeclareMathAlphabet{\mathsfit}{\encodingdefault}{\sfdefault}{m}{sl}
\SetMathAlphabet{\mathsfit}{bold}{\encodingdefault}{\sfdefault}{bx}{n}

\usepackage{hyperref}
\usepackage{url}

\usepackage{mathrsfs}
\usepackage{bm}

\title{Categorical SDEs with Simplex Diffusion}

\author{Pierre H. Richemond\footnotemark[1],\quad Sander Dieleman \& \quad Arnaud Doucet\thanks{ Equal contribution}  \\
DeepMind \\
\texttt{\{richemond, sedielem, arnauddoucet\}@deepmind.com} \\
}

\iclrfinalcopy 
\begin{document}

\maketitle

\begin{abstract}
Diffusion models typically operate in the standard framework of generative modelling by producing continuously-valued datapoints. To this end, they rely on a progressive Gaussian smoothing of the original data distribution, which admits an SDE interpretation involving increments of a standard Brownian motion. However, some applications such as text generation or reinforcement learning might naturally be better served by diffusing categorical-valued data, i.e., lifting the diffusion to a space of probability distributions. To this end, this short theoretical note proposes \emph{simplex diffusion}, a means to directly diffuse datapoints located on an $n$-dimensional probability simplex. We show how this relates to the Dirichlet distribution on the simplex and how the analogous SDE is realized thanks to a multi-dimensional \emph{Cox--Ingersoll--Ross} process (abbreviated as CIR), previously used in economics and mathematical finance. Finally, we make remarks as to the numerical implementation of trajectories of the CIR process, and discuss some limitations of our approach.
\end{abstract}

\section{Introduction and background}

Diffusion models \citep{sohl2015deep, Song2019GenerativeMB, ho2020denoising,song2020score} are a now well-established class of generative models that find applications notably in the image \citep{Dhariwal2021DiffusionMB, Ramesh2022HierarchicalTI, Saharia2022PhotorealisticTD}, video \citep{Singer2022MakeAVideoTG, Villegas2022PhenakiVL, Ho2022ImagenVH}, speech \citep{Jeong2021DiffTTSAD, Huang2022ProDiffPF} domains, and even for molecule generation \citep{Hoogeboom2022EquivariantDF, Corso2022DiffDockDS}. These models proceed as follows. One adds noise progressively to data using a diffusion process to transform the complex data distribution to a simple easy-to-sample distribution. The generative model is obtained by simulating an approximation of the time-reversal of this process. The resulting ``denoising'' process is also a diffusion whose drift depends on the logarithmic gradients of
the noised data densities \citep{Anderson1982ReversetimeDE}, i.e. the Stein \emph{scores}. These scores are estimated using a neural network via score matching
\citep{Hyvarinen:2005a}. In the usual case where Gaussian noise is progressively added to the generative distribution, the score matching objective simply reduces to a least squares denoising term \citep{vincent2011connection} easily amenable to gradient descent. 

In all which precedes, the datapoints are assumed to be vectors taking continuous values. Being able to proceed with diffusion when those datapoints are instead discrete-valued would further widen the applicability domain of diffusion models, in particular to language modelling \citep{Savinov2022StepunrolledDA, Li2022DiffusionLMIC, Wang2022LanguageMV} and even reinforcement learning \citep{Richemond2017OnWR, Janner2022PlanningWD}. We propose a construction of such a discrete diffusion in this short technical note. Our approach consists in directly deriving a tractable stochastic process that operates on the probability simplex itself, lifting traditional diffusion schemes to categorical distributions, rather than relying on auxiliary methods such as binary encoding \citep{Chen2022AnalogBG}. Because of this, we can use simplex diffusion in conjunction with the now standard mathematical machinery of diffusion models, including equivalent ODE formulation, and computation of an evidence lower bound (ELBO). Finally, we also discuss some specific limitations of our approach, namely the issues once encounters in practice when simulating high-dimensional simplices (i.e., for large values of $n$).

\section{Simplex diffusion with the Cox--Ingersoll--Ross process}

We first proceed to recall how one can sample from the Dirichlet distribution on the probability simplex using independent Gamma random variables. Then, we introduce a compatible stochastic process, the Cox--Ingersoll--Ross process.

\subsection{Dirichlet distribution on the simplex}

For a given integer $n \geq 2$, the $n-1$ dimensions probability simplex is the set of $n$-dimensional vectors $\bm{X}$ in $\mathbb{R}^n$ whose components $\bm{X}:=(X_1,...,X_n)$ satisfy $X_i \geq 0$ and $\sum_{i=1}^n X_i=1$. A point on the simplex is hence assimilated to an $n$-way categorical distribution.

The Dirichlet distribution is defined over the simplex as the conjugate prior of the categorical distribution. It is a multivariate, continuous distribution, parametrized by an arbitrary vector of strictly positive scalars $\bm{\alpha}$. The Dirichlet distribution $\mathcal{D}(\bm{\alpha})$ with parameters $\bm{\alpha}:=(\alpha_1,...,\alpha_n)$, where $\alpha_1,...,\alpha_n>0$, has probability density function $f_{\mathcal{D}}$ given (w.r.t. the standard Lebesgue measure on $\mathbb{R}^n$) by
\begin{equation}
    f_{\mathcal{D}}(x_1, \cdots, x_n; \alpha_1, \cdots, \alpha_n) = \frac{1}{Z(\bm{\alpha})} \prod_{i=1}^{n}{x_{i}^{\alpha_i - 1}}
\end{equation}
with $Z(\bm{\alpha})$ a normalizing constant. In particular, the choice $\alpha_i=1$ for all $i$ recovers a uniform distribution over the simplex. In our construction, the Dirichlet distribution plays a role somewhat analogous to that of the Gaussian distribution in standard diffusions - in that it represents the desired stationary distribution of the diffusion process we will build below. Hence, and given its flexibility we focus on it, although other choices of simplex distributions are possible \citep{AitchisonCompositional}.

\textbf{Sampling.} It is well known that sampling from the Dirichlet distribution reduces to a two-step procedure: first, sampling $n$ \emph{independent} Gamma random variables $Y_1 \sim \mathcal{G}(\alpha_1,\beta),...,Y_n \sim \mathcal{G}(\alpha_n,\beta)$ where $\alpha_i$ is their shape parameter, and $\beta$ their common rate parameter. Second, normalizing those random variables to sum to $1$ then yields the Dirichlet-distributed random vector

\begin{equation}
    \bm{X}=\bigg(\frac{Y_1}{\sum_{i=1}^n Y_i},...,\frac{Y_n}{\sum_{i=1}^n Y_i} \bigg) \sim \mathcal{D}(\bm{\alpha}).
\end{equation}

This result holds for any $\beta>0$ so we can use $\beta=1$ specifically. Taking these observations together, we now seek to find an $n$-dimensional stochastic process whose marginal distributions each converge to Gamma laws in the large-time limit. We exhibit such a process below. 

\subsection{The Cox-Ingersoll-Ross process}

The Cox--Ingersoll--Ross (or CIR) process introduced in \cite{cox1985theory} is a popular real-valued diffusion process used in econometrics and quantitative finance, both for yield curve (usually, the instantaneous interest rate) and stochastic equity volatility \citep{Heston1993ACS} modelling. It is an instance of \emph{square-root diffusion} defined by the following SDE in $\theta_t$: for any $\theta_0 \geq 0$ and $a,b,\sigma>0$

\begin{equation}\label{eq:generalCIR}
    \mathrm{d}\theta_t=b(a-\theta_t)\mathrm{d}t+\sigma \sqrt{\theta_t}\mathrm{d}W_t,
\end{equation}

where $(W_t)_{t\geq 0}$ is a standard Brownian motion (or Wiener process). The solution to this SDE exists and is unique \citep{Watanabe1971OnTU}, despite the non-regularity of the square root term near zero. The CIR process is ergodic, almost surely non-negative and admits as invariant limiting distribution the Gamma distribution $\mathcal{G}(2ab/\sigma^2,2b/\sigma^2)$. If $2ab \geq \sigma^2$ and $\theta_0>0$, then the process is strictly positive, pathwise.

For our purpose, we can set $2b=\sigma^2$ so that (\ref{eq:generalCIR}) becomes 
\begin{equation}\label{eq:simplifiedCIR}
    \mathrm{d}\theta_t=b(a-\theta_t)\mathrm{d}t+ \sqrt{2b \theta_t}\mathrm{d}W_t
\end{equation}
and admits the Gamma distribution $\mathcal{G}(a,1)$ as limiting distribution.

\textbf{Conditional mean and variance.} One can readily check that for $t>0$
\begin{equation}
    \mathbb{E}[\theta_t|\theta_0]=\theta_0 \exp(- bt)+ a(1-\exp(-bt))=a+\exp(-bt)(\theta_0-1)
\end{equation}
while 
\begin{align}
    \textup{var}[\theta_t|\theta_0]&=2\theta_0(\exp(-bt)-\exp(-2bt))+a (1-\exp(-bt))^2 \nonumber\\
    &=2\theta_0(\exp(-bt)-\exp(-2bt))+a (1+\exp(-2bt)-2\exp(-bt)) \nonumber\\
    &= a+2\exp(-bt)(\theta_0-a)+\exp(-2t)(a-2\theta_0).
\end{align}

$b$ can be thought of as the parameter governing diffusion speed. As $t \rightarrow \infty$, we have $\mathbb{E}[\theta_t|\theta_0]\rightarrow a$ and $\textup{var}[\theta_t|\theta_0]\rightarrow a$. At any point in time, the drift term in equation \ref{eq:generalCIR} pushes $\theta_t$ back towards its long-term average $a$, a phenomenon known as \emph{mean-reversion}. For this reason $b$ is also indicative of, and sometimes called, the speed of mean-reversion.

\textbf{Density of increments.} The transition density of the CIR process is available in closed-form thanks to Laplace transform techniques \citep{Feller1951TWOSD} and can be sampled from exactly; i.e. we have 
\begin{equation}
    \theta_t|\theta_0 \sim \frac{1-\exp(-bt)}{2}K,\qquad K\sim \chi^2\Big(2a,2\theta_0 \frac{\exp(-bt)}{1-\exp(-bt)}\Big),
\end{equation}
where $\chi^2(\nu,\mu)$ denotes the non-central chi-squared distribution with $\nu$ degrees of freedom and non-centrality parameter $\mu$. We can write this density explicitly as 
\begin{equation}
    f(\theta_t|\theta_0)=c \exp\Big(-c (\theta_0 \exp(-bt)+\theta_t)\Big) \Big(\frac{\theta_t \exp(bt)}{\theta_0}\Big)^{\frac{a-1}{2}}I_{a-1}\Big(2c \sqrt{\theta_0 \theta_t \exp(-bt)}\Big),
\end{equation}
for $c=(1-\exp(-bt))^{-1}$ and $I_{a-1}$ being the modified Bessel function of the first kind of order $a-1$. This closed-form expression for the transition density for the CIR model makes usual denoising score matching techniques applicable, as we'll see below.

\subsection{Simplex diffusion}

\textbf{Simplex SDE.} Our original purpose is to exhibit a diffusion whose marginal distribution, in the large time limit, provides samples from a Dirichlet distribution $\mathcal{D}(\bm{\alpha})$. It follows directly from previous section that this can be achieved by simulating first $n$ independent CIR processes $Y^i$ in parallel, resulting in a process $\bm{Y}_t$ with values in the positive orthant, following (with $\mathrm{d}W^i_t$ the independent increments of a standard $n$-dimensional Brownian motion $\bm{W}_t$, so that $\langle \mathrm{d}W^i_t, \mathrm{d}W^j_t\rangle = \delta_{i,j} t$) :
\begin{equation}\label{eq:simplifiedCIR}
    \mathrm{d}Y^i_t=b(\alpha_i-Y^i_t)\mathrm{d}t+ \sqrt{2b Y^i_t}\mathrm{d}W^i_t
\end{equation}
each $Y^i$ thus having limiting distribution $\mathcal{G}(\alpha_i,1)$. We then consider the normalized, unit-sum vector
\begin{equation}
    \bm{X}_t=\bigg( \frac{Y^1_t}{\sum_{i=1}^n Y^i_t},...,\frac{Y^n_t}{\sum_{i=1}^n Y^i_t}\bigg)
\end{equation}
This normalization projects $\bm{Y}_t$ from the positive orthant to the probability simplex. By construction, we have $\bm{X}_t=(X^1_t,...,X^n_t) \sim \mathcal{D}(\bm{\alpha})$ as $t\rightarrow \infty$. This is our main result, and enables us to perform diffusion towards a vertex of the simplex (a one-hot vector, representing the state of a categorical variable) in the time-reversal process.

Now since this multidimensional SDE retains a standard Brownian increment, both the time-reversal of the SDE, and reformulation as a standard ODE for 'probability flow'-type sampling \citep{song2020score} proceed as usual. We detail those aspects below.

\textbf{Time reversal.} The SDE in equation (\ref{eq:simplifiedCIR}) is of the general (vector) form
\begin{equation}\label{eq:baseSDE}
       \mathrm{d}\bm{Y}_t=\bm{f}_t(t,\bm{Y}_t)\mathrm{d}t+ \bm{G}(t,\bm{Y}_t)\mathrm{d}\bm{W}_t,
\end{equation}
where $\bm{f}(t,\bm{Y}_t)=b( \bm{\alpha} - \bm{Y}_t)$, $\bm{\alpha}=(\alpha_1,...,\alpha_n)^\top$ and $\bm{G}(t,\bm{Y}_t)=\sqrt{2b} \cdot \textup{diag}(\sqrt{Y_t^1},...,\sqrt{Y^n_t})$. Let us also introduce further notation: $p_t=\textup{Law}(\bm{Y}_t)$, the law of the probability density function of $\bm{Y}_t$, and $\bm{\Sigma}(t,\bm{Y}_t)=\bm{G}(t,\bm{Y}_t)\bm{G}(t,\bm{Y}_t)^{\top}=2b \cdot \textup{diag}(\bm{Y}_t)$.

The \emph{time reversal} \citep{Anderson1982ReversetimeDE, haussmann1986time} of the multidimensional CIR given by equation (\ref{eq:simplifiedCIR}) is the process $(\bm{Z}_t)_{t\in [0,T]}$ such that $\bm{Z}_t=\bm{Y}_{T-t}$ satisfies
\begin{align}
    \mathrm{d}\bm{Z}_t  &=\big[-b(\bm{\alpha}-\bm{Z}_t)+2b \cdot \textup{diag}(\bm{Z}_t)~\nabla_{\bm{Z}} \log p_{T-t}(\bm{Z}_t)+2b \textbf{1} \big]\mathrm{d}t+ \bm{G}(T-t,\bm{Z}_t)\mathrm{d}\bm{W}_t \label{eq:reversals1} 
\end{align}
with $\bm{Z}_0 \sim p_T$. In practice, we will approximate this time reversal by the diffusion
\begin{equation}
    \mathrm{d}\bm{Z}_t =\big[-b(\bm{\alpha}-\bm{Z}_t)+2b \cdot \textup{diag}(\bm{Z}_t)~ \bm{s}_{T-t}(\bm{Z}_t)+ 2b \textbf{1} \big]\mathrm{d}t+ \bm{G}(T-t,\bm{Z}_t)\mathrm{d}\bm{W}_t, \label{eq:reversals2}
\end{equation}
with $\bm{Z}_0 \sim p_\textup{ref}$ where $p_\textup{ref}(z^1,...,z^n)=\prod_{i=1}^n\mathcal{G}(z^i;\alpha_i,1)$. Here $\bm{s}_t(\bm{x})$ is a neural score network approximating $\nabla_{\bm{x}} \log p_t(\bm{x})$.

\textbf{Max likelihood training.} This form also lends itself to computation of an evidence lower bound. Maximizing the likelihood of the data is equivalent to minimizing the KL divergence between the terminal time marginal induced by our SDE and the data distribution, which we compute exactly as in \cite[Section 4]{song2021maximum}.
Let $\mathcal{P}$ and  $\mathcal{\hat{P}}$ the path measures corresponding respectively to equations (\ref{eq:reversals1}) and (\ref{eq:reversals2}). Then by Girsanov theorem, the KL-divergence $\textup{KL}(\mathcal{P}||\mathcal{\hat{P}})$ satisfies
\begin{equation}
    \textup{KL}(\mathcal{P}||\mathcal{\hat{P}})=\textup{KL}(p_T||p_\textup{ref})+ \Delta I
\end{equation}
with the integral difference $\Delta I$ given by
\begin{align*}
     \Delta I &= \frac{1}{2}\mathbb{E}_{\mathcal{P}}\left[\int_0^T \Big\| \bm{\Sigma}(T-t,\bm{Z}_t) \nabla_{\bm{Z}} \log p_{T-t}(\bm{Z}_t)-\bm{\Sigma}(T-t,\bm{Z}_t) \bm{s}_{T-t}(\bm{Z}_t) \Big\|^2_{\bm{\Sigma}^{-1}(T-t,\bm{Z}_t)} \mathrm{d}t \right]\\
    &=\frac{1}{2}\mathbb{E}_{\mathcal{P}}\left[\int_0^T \Big\|\nabla_{\bm{Y}} \log p_{t}(\bm{Y}_t)-\bm{s}_{t}(\bm{Y}_t) \Big\|^2_{\bm{\Sigma}(t,\bm{Y}_t)} \mathrm{d}t \right],
\end{align*}
where we use the notation $||\bm{x}||_{\bm{A}}=\bm{x}^{\top}\bm{A}\bm{x}$.
Now thanks to the denoising score matching trick, we get that, up to the additive constant (w.r.t. optimization) term $\textup{KL}(p_T||p_\textup{ref})$,
\begin{align*}
    \textup{KL}(\mathcal{P}||\mathcal{\hat{P}}) &\equiv \frac{1}{2}\mathbb{E}_{\mathcal{P}}\left[\int_0^T \Big\|\nabla_{\bm{Y}} \log p_{t|0}(\bm{Y}_t|\bm{Y}_0)-\bm{s}_{t}(\bm{Y}_t) \Big\|^2_{\bm{\Sigma}(t,\bm{Y}_t)} \mathrm{d}t \right] \nonumber\\
    &\equiv \mathbb{E}_{\mathcal{P}}\left[\int_0^T b \Big(\nabla_{\bm{Y}} \log p_{t|0}(\bm{Y}_t|\bm{Y}_0)-\bm{s}_{t}(\bm{Y}_t) \Big)^{\top}\textup{diag}(\bm{Y}_t)\Big(\nabla_{\bm{Y}} \log p_{t|0}(\bm{Y}_t|\bm{Y}_0)-\bm{s}_{t}(\bm{Y}_t) \Big) \mathrm{d}t \right]
\end{align*}

\textbf{ODE formulation for sampling.} The ODE formulation consists in finding an ODE
\begin{equation}\label{eq:probaflowODE}
       \frac{\mathrm{d}\bm{Y}_t}{\mathrm{d}t}=\tilde{\bm{f}}_t(t,\bm{Y}_t)
\end{equation}
that admits the same temporal marginals as the solution of equation (\ref{eq:simplifiedCIR}). Using the formulation in \cite{song2020score}, or simply by applying Ito's lemma, one gets:
\begin{equation}
       \tilde{\bm{f}}_t(t,\bm{Y}_t)= \bm{f}_t(t,\bm{Y}_t)-\frac{1}{2}\nabla \cdot \bm{\Sigma}(t,\bm{Y}_t)-\frac{1}{2} \bm{\Sigma}(t,\bm{Y}_t) \nabla_{\bm{Y}} \log p_t(\bm{Y}_t),
\end{equation}
which in our case results in

\begin{equation}
    \frac{\mathrm{d}\bm{Y}_t}{\mathrm{d}t} =b(\bm{\alpha} -\textbf{1} -\bm{Y}_t - \textup{diag}(\bm{Y}_t) \nabla_{\bm{Y}} \log p_t(\bm{Y}_t))
\end{equation}

This highlights another benefit of the ODE formulation: we can simulate the ODE in the log-domain and get an equation of the form $\frac{\mathrm{d} \log \bm{y}}{\mathrm{d} t} = -b (1 + \nabla_{\bm{y}} \log p_{t}(\bm{y}) )$, to promote numerical stability.

\textbf{Remarks on numerical simulation.} The CIR process has been extensively used and studied within Monte Carlo methods \citep{Glasserman2003MonteCM} in quantitative finance. Care must be taken in simulating its trajectories; this can typically require an additional scalar $\epsilon$ stabilization parameter inside of the square-root diffusion term in equation \ref{eq:simplifiedCIR} in order to avoid path termination due to discretization error. Another avenue is to observe that under specific conditions on their parameters, the sum of independent, squared Ornstein--Uhlenbeck processes is identical in law to a CIR process \citep{Jamshidian1995ASC}; this observation relates to Bessel processes \citep{Revuz1990ContinuousMA}. This enables substituting a single CIR path for multiple Ornstein-Uhlenbeck paths, trading off stability for computation.

\textbf{Limitations.} We might want to use our approach on very high dimensional simplices in order to simulate \emph{one-of-many} categoricals - for instance, when modelling language tokens over a sizeable vocabulary, or in the case of a large action-space policy. This comes with practical issues, chief amongst those being the potential presence of outliers in the categorical distribution. When we draw a sample from the transition density of the CIR process for a given $t$, we can determine the rank of the ground truth token in the resulting (unnormalized) vector. We observed in practice that the distribution of that rank - whose closed form law involves large, and possibly intractable integrals - is extremely heavy-tailed. Informally, this can lead to noisy results. We found empirically this phenomenon to be particularly relevant in high dimensions.

Finally, we note that while the interpretation of noisy vectors as unnormalized probability distributions via a Dirichlet prior is useful to build intuition, it is not rigorous. When one considers the posterior distribution at token level $p_t(\bm{x}_0 | \bm{x}_t)$, where $\bm{x}_0$ is a one-hot vector representing a token, and $\bm{x}_t$ is the noisy unnormalized probability input vector, we can apply Bayes' rule and get 
\begin{equation}
p_t(\bm{x}_0 | \bm{x}_t) = \frac{p_t(\bm{x}_t|\bm{x}_0) p(\bm{x}_0)}{\sum_{\bm{x}_0} p_t(\bm{x}_t|\bm{x}_0) p(\bm{x}_0)}
\end{equation}
thus showing that $p(\bm{x}_0|\bm{x}_t)$ is actually nonlinear in $\bm{x}_t$.

\textbf{Related and alternative approaches.} The Cox--Ingersoll--Ross process is seldom used in machine learning. Similar derivations to ours nonetheless previously appeared in \cite{Baker2018LargeScaleSS}, where a CIR process is also used to approximate a Dirichlet distribution, but in a Bayesian inference context, with the very different purpose of obviating discretization error in stochastic gradient MCMC \citep{Welling2011BayesianLV, Ma2015ACR}. Other stochastic processes than the CIR can be built that admit the Dirichlet distribution as a limiting distribution. \cite{evans2003diffusions} considers functions of the components of a  multivariate Brownian motion running on a hypersphere. When those functions are all identically a squaring, by construction the squared components sum to $1$ and can thus represent a categorical probability vector. In that setting the invariant distribution of the squared-components vector is proven to be symmetric Dirichlet with parameter $1/2$. Unlike ours, that approach is however not fully compatible with standard diffusion score matching, since the transition density of the Brownian motion on the sphere is to our knowledge not known in closed form - it is merely possible to sample from \citep{Mijatovic2020AnAF}. Other choices than a Dirichlet limiting distribution are also possible, even as it represents a reasonable and flexible prior family; \cite{AitchisonCompositional} proposes a generic \emph{log-ratio transform} projecting unconstrained, multivariate distributions defined on $\mathbb{R}^n$ onto the simplex. Separately, \cite{Lafferty2005DiffusionKO} perform an asymptotic expansion of the heat kernel on statistical manifolds (including an approximation of the simplex), with application to the multinomial family of distributions towards text classification.

\section{Conclusion}

We have introduced \emph{simplex diffusion}, a simple method that uses a multi-dimensional Cox-Ingersoll-Ross process, via a unit-sum normalization of its time marginals, to diffuse categorical distributions directly on the probability simplex. Our approach is tractable and compatible with the tools of standard stochastic calculus central to diffusion models. Further research will involve operationalizing and evaluating deep learning models that leverage this principle.

\nocite{evans2003diffusions, Baker2018LargeScaleSS}

\bibliography{iclr2023_conference}
\bibliographystyle{iclr2023_conference}

\end{document}